\documentclass{llncs}
\usepackage{makeidx}
\usepackage{tabularx,booktabs}
\usepackage{graphicx}
\usepackage{color,soul}
\usepackage{amsmath}
\usepackage{wrapfig}

\begin{document}

\title{Simultaneous Segmentation and Classification of Bone Surfaces from Ultrasound Using a Multi-feature Guided CNN}
%
%
\author{Puyang Wang\inst{1}, Vishal M. Patel\inst{1} \and  Ilker Hacihaliloglu\inst{2,3}}
%
%
%
\institute{Department of Electrical and Computer Engineering, Rutgers University, Piscataway, NJ\\
\and
Department of Biomedical Engineering, Rutgers University, Piscataway, USA
\and
Department of Radiology, Rutger Robert Wood Johnson Medical School, New Brunswick, NJ, USA}

\maketitle              
\vspace*{-4mm}
\begin{abstract}
Various imaging artifacts, low signal-to-noise ratio, and bone surfaces
appearing several millimeters in thickness have hindered the success
of ultrasound (US) guided computer assisted orthopedic surgery
procedures. In this work, a multi-feature guided convolutional neural network (CNN) architecture is proposed for simultaneous enhancement, segmentation, and classification of bone surfaces from US data. The proposed CNN consists of two main parts: a pre-enhancing net, that takes the concatenation of B-mode US scan and three filtered image features for the enhancement of bone surfaces, and a modified U-net with a classification layer. The proposed method was validated on 650 in vivo US scans collected using two US machines, by scanning knee, femur, distal radius and tibia bones. Validation, against expert annotation, achieved statistically significant improvements in segmentation of bone surfaces compared to state-of-the-art.
\end{abstract}
\vspace*{-10mm}
\section{Introduction}
\label{sec:intro}
In order to provide a radiation-free, real-time, cost effective imaging alternative, for intra-operative fluoroscopy, special attention has been given to incorporate ultrasound (US) into computer assisted orthopedic surgery (CAOS) procedures \cite{hacihaliloglu2017ultrasound}. However, problems such as high levels of noise, imaging artifacts, limited field of view and bone boundaries appearing several millimeters (mm) in thickness have hindered the wide spread adaptability of US-guided CAOS systems. This has resulted in the development of automated bone segmentation and enhancement methods \cite{hacihaliloglu2017ultrasound}. Accurate and robust segmentation is important for improved guidance in US-based CAOS procedures.



In discussing state-of-the-art we will limit ourselves to approaches that fit directly within the context of the proposed deep learning-based method. A detailed review of image processing methods based on the extraction of image intensity and phase information can be found in \cite{hacihaliloglu2017ultrasound}. In \cite{salehi2017precise}, U-net architecture, originally proposed in \cite{ronneberger2015u}, was investigated for processing in vivo femur, tibia and pelvis bone surfaces. Bone localization accuracy was not assessed but 0.87 precision and recall rates were reported. In \cite{baka2017ultrasound}, a modified version of the CNN proposed in \cite{ronneberger2015u} was used for localizing vertebra bone surfaces. 
Despite the fact that methods based on deep learning produce robust and accurate results, the success rate is dependent on: (1) number of US scans used for training, (2) quality of the collected US data for testing \cite{baka2017ultrasound}.


In this paper, we  propose a novel  neural network architecture for simultaneous bone surface enhancement, segmentation and classification from US data. Our proposed network accommodates a bone surface enhancement network which takes a concatenation of B-mode US scan, local phase-based enhanced bone images, and signal transmission-based bone shadow enhanced image as input and outputs a new US scan in which only bone surface is enhanced. We show that the bone surface enhancement network, referred to as pre-enhancing (\textit{PE}), improves robustness and accuracy of bone surface localization since it creates an image where the bone surface information is more dominant. As a second contribution, a deep-learning bone surface segmentation framework for US image, named classification U-net, \textit{cU-net} for short, is proposed. Although \textit{cU-net} shares the same basic structure with U-net \cite{ronneberger2015u}, it is fundamentally different in terms of designed output. Unlike U-net, \textit{cU-net} is capable of identifying bone type and segmenting bone surface area in US image simultaneously. The bone type classification is implemented by feeding part of the features in U-net to a sequence of fully-connected layers followed by a softmax layer. To take the advantages of both \textit{PE} and \textit{cU-net}, we propose a framework that can adaptively balance the trade-off between accuracy and running-time by combining \textit{PE} and \textit{cU-net}. 
\vspace*{-4mm}

\section {Proposed Method}

\begin{figure*}[htp!]
	\centering
	\includegraphics[width=110mm]{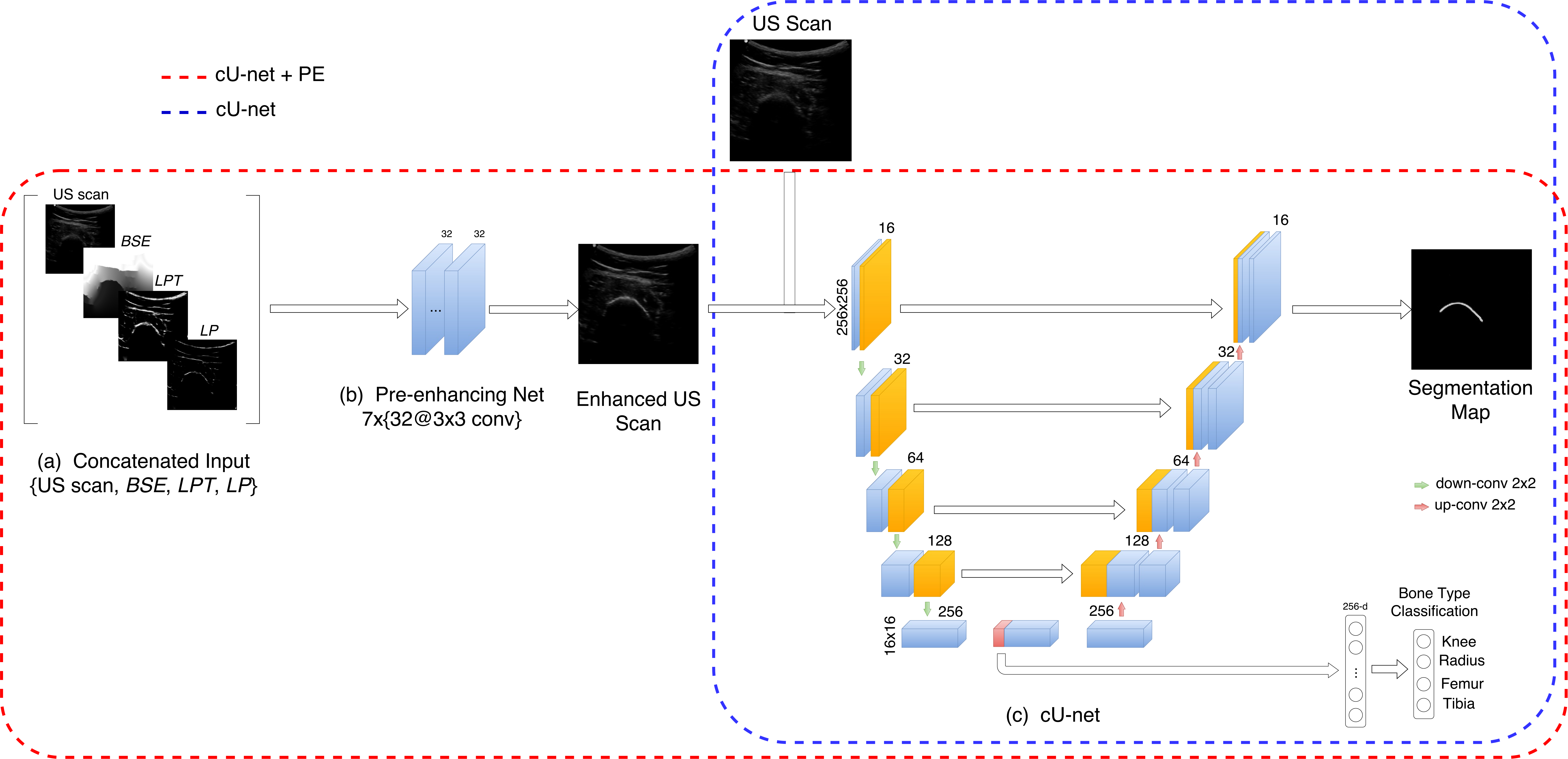}
	\vskip -8pt
	\caption{Overview of the proposed simultaneous enhancement, segmentation and classification network.}
	\label{fig:overview}
	\vspace*{-5mm}
\end{figure*}

Fig.\ref{fig:overview} gives an overview of the proposed joint bone enhancement, segmentation and classification framework. Incorporating pre-enhancing net, \textit{cU-net+PE}, into the proposed framework is expected to produce more accurate results than using only \textit{cU-net}. However, because of the computation of the additional input features and convolution layers, \textit{cU-net+PE} requires more running time. Therefore, the proposed framework can be configured for both (i)real-time application using only \textit{cU-net}, and (ii)off-line application using \textit{cU-net+PE} for different clinical purposes. In the next section, we explain how the various filtered images are extracted. 

\noindent\textbf{2.1 Enhancement of Bone Surface and Bone Shadow Information}
\newline Different from using only B-mode US scan as input, the proposed pre-enhancing network, that enhances bone surface, takes the concatenation of B-mode US scan ($US(x,y)$) and three filtered image features which are obtained as follows:

\noindent\textbf{Local Phase Tensor Image ($LPT (x,y)$):} $LPT(x,y)$ image is computed by defining odd and even filter responses using \cite{hacihaliloglu2014local}: 
\setlength{\arraycolsep}{2pt}
\begin{eqnarray}
T_{even}=\left[\textit{\textbf{H}}(US_{DB}(x,y))\right]\left[\textit{\textbf{H}}(US_{DB}(x,y))\right ]^{T} \text{,}\\ 
T_{odd}=-0.5\times(\left[\nabla US_{DB}(x,y)\right]\left[\nabla\nabla^2 US_{DB}(x,y)\right]^{T}+\nonumber\\
\left[\nabla\nabla^2 US_{DB}(x,y)\right]\left[\nabla US_{DB}(x,y)\right]^{T}) \text{.}\nonumber
\end{eqnarray}
\setlength{\arraycolsep}{5pt}
\noindent Here $T_{even} $  and $ T_{odd} $ represent the symmetric and asymmetric features of $US(x,y)$. $\textbf{H}$, $\nabla $ and $\nabla^{2}$ represent the Hessian, Gradient and Laplacian operations, respectively. In order to improve the enhancement of bone surfaces located deeper in the image and mask out soft tissue interfaces close to the transducer, $US(x,y)$ image is masked with a distance map and band-pass filtered using Log-Gabor filter\cite{hacihaliloglu2014local}. The resulting image, from this operation, is represented as $US_{DB}(x,y)$. The final $LPT(x,y)$ image is obtained using: $ LPT(x,y)= \sqrt {T^{2}_{even} + T^{2}_{odd} } \times cos(\phi) $, where $\phi$ represents instantaneous phase obtained from the symmetric and asymmetric feature responses, respectively \cite{hacihaliloglu2014local}.\vspace*{1mm}

\noindent\textbf{Local Phase Bone Image ($LP(x,y)$):} $LP(x,y)$ image is computed using: $LP(x,y)=LPT(x,y) \times LPE(x,y) \times LwPA(x,y)$, where $LPE(x,y)$ and $LwPA(x,y)$ represent the local phase energy and local weighted mean phase angle image features, respectively. These two features are computed using monogenic signal theory as \cite{hacihaliloglu2017enhancement}: $LPE(x,y)=\sum_{sc}|US_{M1}(x,y)|-
\sqrt{US_{M2}^{2}(x,y)+US_{M2}^{3}(x,y)},$
\begin{equation}
LwPA(x,y)=\arctan{\dfrac{\sum_{sc}US_{M1}(x,y)}{\sqrt{\sum_{sc}US_{M1}^{2}+\sum_{sc}US_{M2}^{2}(x,y)}}},
\end{equation}\newline
\noindent where $US_{M1},US_{M2},US_{M3}$ represent the three different components of monogenic signal image ($US_{M}(x,y)$) calculated from $LPT(x,y)$ image using Riesz filter \cite{hacihaliloglu2017enhancement} and $sc$ represents the number of filter scales. \vspace*{1mm}

\noindent\textbf{Bone Shadow Enhanced Image ($BSE(x,y)$):} $BSE(x,y)$ image is computed by modeling the interaction of the US signal within the tissue as scattering and attenuation information using \cite{hacihaliloglu2017enhancement}:
\begin{equation}
BSE(x,y)=[(CM_{LP}(x,y)-\rho)/[max(US_{A}(x,y),\epsilon)]^\delta]+\rho,
\end{equation}
\noindent where $CM_{LP}(x,y)$ is the confidence map image obtained by modeling the propagation of US signal inside the tissue taking into account bone features present in  $LP(x,y)$ image \cite{hacihaliloglu2017enhancement}. $US_{A}(x,y)$, maximizes the visibility of high intensity bone features inside a local region and satisfies the constraint that the mean intensity of the local region is less than the echogenicity of the tissue confining the bone \cite{hacihaliloglu2017enhancement}. Tissue attenuation coefficient is represented with $\delta$. $\rho$ is a constant related to tissue echogenicity confining the bone surface, and $\epsilon$ is a small constant used to avoid division by zero \cite{hacihaliloglu2017enhancement}. \vspace*{2mm}  

\begin{figure*}[htp!]
	\centering

	\includegraphics[width=90mm]{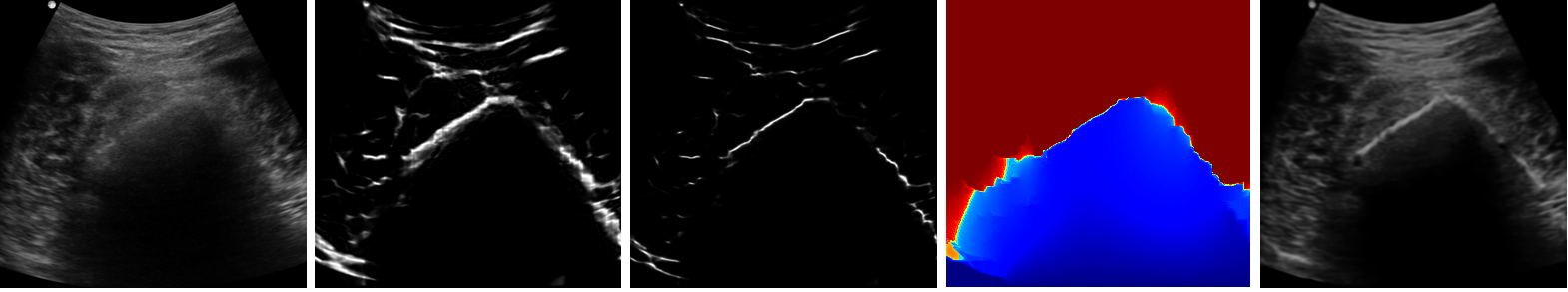}\\
	\vskip -8pt\caption{From left to right: B-mode US scan, \textit{LPT}, \textit{LP}, \textit{BSE}, bone-enhanced US scan.}
	\label{fig:spine}
	\vskip -8pt
	\vspace*{-4mm}
\end{figure*}

\noindent\textbf{2.2 Pre-enhancing Network (\textit{PE}}) 
\newline A simple and intuitive way to view the three extracted feature images is viewing them as an input feature map of a CNN. Each feature map provides different local information of bone surface in an US scan. In deep learning, if a network is trained on a dataset of a specific distribution and is tested on a dataset that follows another distribution, the performance usually degrades significantly. In the context of bone segmentation, different US machines with different settings or different orientation of the transducer will lead to scans that have different image characteristics. The main advantage of multi-feature guided CNN is that filtered features can bring the US scan to a common domain independent of the image acquisition device. Hence, the bone surface in a US scan appears more dominant after the multi-feature guided pre-enhancing net regardless of different US image acquisition settings (Fig.\ref{fig:spine}). \vspace*{-6mm} 

The input data consists of a $4\times256\times256$ matrix, i.e., each channel consists of a $256\times256$ image. The pre-enhancing network (\textit{PE}) contains seven convolutional layers with 32 feature maps and one with single feature map (Fig.\ref{fig:overview} (b)). To balance the trade-off between the large receptive field, which can acquire more semantic spatial information and the increase in the number of parameters, we set the convolution kernel size to be $3\times3$ with zero-padding of size 1. The batch normalization (BN) \cite{batch_normalization} and rectified linear units (ReLU) are attached to every convolutional layer except the last one for faster training and non-linearity.  Finally, the last layer is a Sigmoid function that transforms the single feature map to visible image of values between $[0,1]$. Next we explain the proposed simultaneous segmentation and classification method. \vspace*{2mm}

\noindent\textbf{2.3 Joint Learning of Classification and Segmentation}
\newline Although U-net has been widely used in many segmentation problems in the field of biomedical imaging, it lacks the capability of classifying medical images. Inspired by the observation that the contracting path of U-net shares similar structure with many image classification networks, such as AlexNet \cite{alexnet}and VGG net \cite{vgg}, we propose a classification U-net (\textit{cU-net}) that can jointly learn to classify and segment images. The network structure is shown in Fig.\ref{fig:overview} (c). 

While our proposed \textit{cU-net} is structurally similar to U-net, three key difference of the proposed cU-net from U-net are as follows:
\begin{enumerate}
  \item The MaxPooling layers and the convolutional layers in the contracting path are replaced by the convolutional layers with stride two. The stride of convolution defines the step size of the kernel when traversing the image. While its default is usually 1, we use a stride of 2 for downsampling an image similar to MaxPooling. Compared to MaxPooling, strided convolution can be regarded as parameterized downsampling that preserves positional information and are easy to reverse. 
  \item Different from \cite{simultaneous}, for the purpose of enabling U-net to classify images, we take only part of the feature maps at the last convolution layer of the contracting path (left side) and expand it as a feature vector. The resulting feature vector is input to a classifier that consists of one fully-connected layer with a final 4-way softmax layer. 
  \item To further accelerate the training process and improve the generalization ability of the network, we adopt BN and add it before every ReLU layers. By reducing the internal covariance shift of features, the batch normalization can lead to faster learning and higher overall accuracy. 
\end{enumerate}

Apart from the above two major differences,  one minor difference is the number of starting feature maps. We reduce the number of starting feature maps from 32 to 16. Overall, the proposed \textit{cU-net} consists of the repeated application of one $3\times3$ convolution (zero-padded convolution), each followed by BN and ReLU, and a $2\times2$ strided convolution with stride 2 (down-conv) for downsampling. At each downsampling step, we double the number of feature maps. Every step in the expansive path consists of an upsampling of the feature map followed by a dilated $2\times2$ convolution (up-conv) that halves the number of feature maps, a concatenation with the corresponding feature map from the contracting path, and one $3\times3$ convolution followed by BN and ReLU.
\vspace*{2mm}

\noindent\textbf{2.4 Data Acquisition and Training}

\noindent After obtaining the institutional review board (IRB) approval, a total of 519 different US images, from 17 healthy volunteers, were collected using SonixTouch US machine (Analogic Corporation, Peabody, MA, USA). The scanned anatomical bone surfaces included knee, femur, radius, and tibia. Additional 131 US scans were collected from two subjects using a hand-held wireless US system (Clarius C3, Clarius Mobile Health Corporation, BC, Canada). All the collected data was annotated by an expert ultrasonographer in the preprocessing stage. Local phase images and bone shadow enhanced images were obtained using the filter parameters defined in \cite{hacihaliloglu2017enhancement}. For the ground truth labels we dilated the ground truth contours to a width of 1 mm. 

We apply a random split of US images from SonixTouch in training (80\%) and testing (20\%) sets. The training set consists of a total of 415 images obtained from SonixTouch only. The rest 104 images from SonixTouch and all 131 images from Clarius C3 were used for testing. We also made sure that during the random split of the SonixTouch dataset the training and testing data did not include the same patient scans. Experiments are carried out three times on random training-testing splits and average results are reported.  For training both \textit{cU-net} and pre-enhancing net (\textit{PE}), we adapt a 2-step training phase. In a total of 30,000 training iterations, the first 10,000 iterations were only performed on \textit{cU-net} and we jointly train the \textit{cU-net} and pre-enhancing net for another 20,000 iterations. We used cross entropy loss for both segmentation and classification tasks of \textit{cU-net}. As for the pre-enhancing net, to force the network only enhance bone surfaces, we used Euclidean distance between output and input as the loss. ADAM stochastic optimization \cite{adam_opt} with batch size of 16 and a learning rate of 0.0002 are used for learning the weights.

For the experimental evaluation and comparison, we selected two reference methods: original U-net \cite{ronneberger2015u} and modified U-net for bone segmentation \cite{baka2017ultrasound} (denoted as $TMI$). For the proposed method, we included two configurations: \textit{cU-net+PE} and \textit{cU-net}, where \textit{cU-net} is the trained model without pre-enhancing net (PE). To further validate the effectiveness of \textit{cU-net} and \textit{PE}, \textit{U-net+PE} (\textit{U-net} trained with enhanced images) and \textit{U-net} trained using same input image features as \textit{PE}  (denoted as \textit{U-net2}) were added to the comparison. All these methods were implemented and evaluated on segmenting several bone surfaces including knee, femur, radius, and tibia. To localize the bone surface, we threshold the estimated probability segmentation map and use the center pixels along each scanline as a single bone surface. The quality of the localization was evaluated by computing average Euclidean distance (AED) between the two surfaces. Apart from AED, we also evaluated the bone segmentation methods with regards to recall, precision, and their harmonic mean, the F-score. Since manual ground truths cannot be regarded as absolute gold standard, true positive are defined as detected bone surface points that are maximum 0.9 mm away from the manual ground truth.
\vspace*{-4mm}

\section{Experimental Results}
\begin{figure*}[htp!]
	\centering
	\includegraphics[width=18mm]{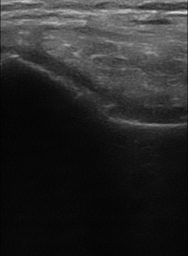}
	\includegraphics[width=18mm]{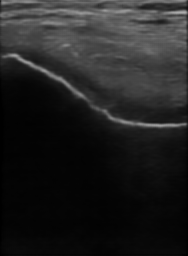}
	\includegraphics[width=18mm]{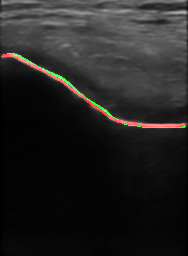}
	\includegraphics[width=18mm]{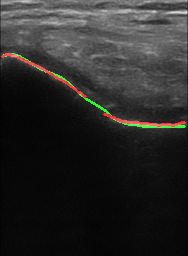}
	\includegraphics[width=18mm]{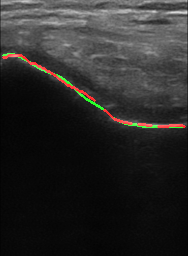}\\
	\vskip -5pt
	{\tiny \hspace{16em}0.99/0.99/0.99 \hspace{1.5em}0.92/0.99/0.95 \hspace{1.5em}0.87/0.98/0.92}\\
	\includegraphics[width=18mm]{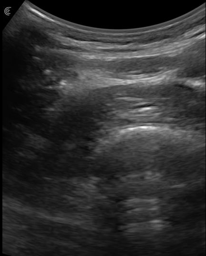}
	\includegraphics[width=18mm]{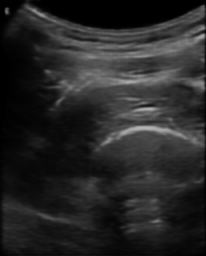}
	\includegraphics[width=18mm]{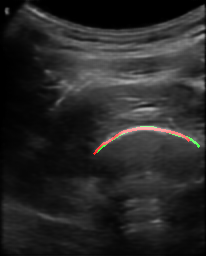}
	\includegraphics[width=18mm]{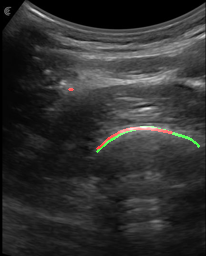}
	\includegraphics[width=18mm]{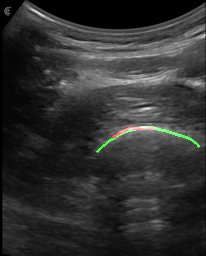}\\
	\vskip -5pt
	{\tiny\hspace{15em} 0.91/0.98/0.94 \hspace{1.5em}0.73/0.95/0.83 \hspace{1.5em}0.68/1/0.81}\\
	\vskip -8pt\caption{From left to right column: B-mode US scans, \textit{PE}, \textit{cU-net+PE}, U-net\cite{ronneberger2015u}, TMI\cite{baka2017ultrasound}. Green represents manual expert segmentation and red is obtained using corresponding algorithms. Recall/Precision/F-score are shown under segmentation results.}
	\label{fig:sample}
\vspace*{-6mm}
\end{figure*}

The AED results (mean$\pm$ std) in Table \ref{tab:results} show that the proposed \textit{cU-net+PE} outperforms other methods on test scans obtained from both US machines. Note that training set only contains images from one specific US machine (SonixTouch) while testing is performed on both. A further paired t-test between \textit{cU-net+PE} and U-net at a 5\% significance level with \textit{p-value} of 0.0014 clearly indicates that the improvements of our method are statistically significant. The \textit{p-values} for the remaining comparisons were also $<0.05$ proving the achieved significance. 
The average recall and precision rates as well as F-scores are reported in Table \ref{tab:results}. Although our method is not performing the best in term of average precision, the more practical measurement for detection tasks, F-score, shows the superiority of our method on bone detection performance. Further experiments of \textit{U-net+PE} and \textit{U-net2} yield 0.949/0.876 and 0.941/0.856 in term of F-score on both US machines. From the fact that \textit{cU-net+PE} $>$ \textit{U-net+PE} $>$ \textit{U-net2}, the proposed \textit{cU-net} and \textit{PE} are shown to improve the segmentation result independently. Qualitative results in Fig.\ref{fig:sample} show that \textit{TMI} method achieves high precision but low recall due to missing bone boundaries which is more important for our clinical application. It can be observed that quantitative results are consistent with the visual results. Average computational time for bone surface and shadow enhancement was 2 seconds (MATLAB implementation). 

\begin{table*}
	
	\caption {AED, 95\% confidence level (CL), recall, precision, and F-scores for the proposed and state of the art methods.}
	
	\resizebox{\textwidth}{!}{%
		
		\centering
		\begin{tabular}{c c c c c c c c c}
			
			\cmidrule{1-9}
			& \multicolumn{4}{c}{SonixTouch}  & \multicolumn{4}{c}{Clarius C3}\\
			\cmidrule(lr){2-5} \cmidrule(l){6-9} 
			 & \textit{cU-net+PE} & \textit{cU-net}  & U-net\cite{ronneberger2015u} & TMI\cite{baka2017ultrasound}  & \textit{cU-net+PE} & \textit{cU-net}  & U-net\cite{ronneberger2015u} & TMI\cite{baka2017ultrasound} \\
			 AED &\textbf{0.246$\pm$0.101}&0.338$\pm$0.158&0.389$\pm$0.221&0.399$\pm$0.201&\textbf{0.368$\pm$0.237}&0.544$\pm$0.876&1.141$\pm$1.665&0.644$\pm$2.656\\
			 95\%CL& \textbf{0.267}&0.371&0.435&0.440&\textbf{0.409}&0.696&1.429&1.103\\
			 Recall& \textbf{0.97}&0.948&0.929&0.891&\textbf{0.873}&0.795&0.673&0.758\\
			 Precision& 0.965&0.943&0.930&\textbf{0.963}&0.94&0.923&0.907&\textbf{0.961}\\
			 F-score& \textbf{0.968}&0.945&0.930&0.926&\textbf{0.906}&0.855&0.773&0.847\\
			 \cmidrule{1-9}
			
		\end{tabular}
	}
	\label{tab:results}
	\vspace*{-2mm}
\end{table*}

Moreover, we evaluate the classification performance of the proposed \textit{cU-net} by calculating classification errors on four different anatomical bone types. The proposed classification U-net, \textit{cU-net}, is near perfect in classifying bones for US images of SonixTouch ultrasound machine with an overall classification error of 0.001. However, the classification errors increase significantly to 0.389 when \textit{cU-net} is tested on test images of Clarius C3 machine. We believe it is because of the imbalanced dataset and dataset bias since the training set only contains 3 tibia images and no images from Clarius C3 machine. Furthermore, Clarius C3 machine is a convex array transducer and is not suitable for imaging bone surfaces located close to the transducer surface which was the case for imaging distal radius and tibia bones. Due to suboptimal transducer and imaging extracted features were not representative of the actual anatomical surfaces.  


%
%

\vspace*{-4mm}
\section{Conclusion}
We have presented a multi-feature guided CNN for simultaneous enhancement, segmentation and classification of bone surfaces from US data. 
To the best of our knowledge this is the first study proposing these tasks simultaneously in the context of bone US imaging. Validation studies achieve a 44\% and 27\% improvement in overall AED errors over the state-of-the-art methods reported in \cite{baka2017ultrasound} and \cite{ronneberger2015u} respectively. In the experiments, our method yields more accurate and complete segmentation even under not only difficult imaging conditions but also different imaging settings compared to state-of-the-art. In this study the classification task involved the identification of bone types. However, this can be changed to identify US scan planes as well. Correct scan plane identification is an important task for spine imaging in the context of pedicle screw insertion and pain management. One of the main drawbacks of the proposed framework is the long computation time required to calculate the various phase image features. However, the proposed \textit{cU-net} is independent of the \textit{cU-net+PE}. Therefore, for real-time applications initial bone surface extraction can be performed using \textit{cU-net} and updated during a second iteration using \textit{cU-net+PE}. Future work will involve extensive clinical validation, real-time implementation of phase filtering, and incorporation of the extracted bone surfaces into a registration method.
\vspace*{2mm}

\noindent\textbf{Acknowledgement:} This work was supported in part by 2017 North American Spine Society Young Investigator Award.
\vspace*{-3mm}
%
%
\bibliographystyle{splncs}
\bibliography{sample}

\begin{thebibliography}{10}

\bibitem{hacihaliloglu2017ultrasound}
Hacihaliloglu, I.:
\newblock Ultrasound imaging and segmentation of bone surfaces: A review.
\newblock Technology (2017)  1--7

\bibitem{salehi2017precise}
Salehi, M., Prevost, R., Moctezuma, J.L., Navab, N., Wein, W.:
\newblock Precise ultrasound bone registration with learning-based segmentation
  and speed of sound calibration.
\newblock In: International Conference on Medical Image Computing and
  Computer-Assisted Intervention, Springer (2017)  682--690

\bibitem{ronneberger2015u}
Ronneberger, O., Fischer, P., Brox, T.:
\newblock U-net: Convolutional networks for biomedical image segmentation.
\newblock In: International Conference on Medical image computing and
  computer-assisted intervention, Springer (2015)  234--241

\bibitem{baka2017ultrasound}
Baka, N., Leenstra, S., van Walsum, T.:
\newblock Ultrasound aided vertebral level localization for lumbar surgery.
\newblock IEEE transactions on medical imaging \textbf{36}(10) (2017)
  2138--2147

\bibitem{hacihaliloglu2014local}
Hacihaliloglu, I., Rasoulian, A., Rohling, R.N., Abolmaesumi, P.:
\newblock Local phase tensor features for 3-d ultrasound to statistical shape+
  pose spine model registration.
\newblock IEEE transactions on medical imaging \textbf{33}(11) (2014)
  2167--2179

\bibitem{hacihaliloglu2017enhancement}
Hacihaliloglu, I.:
\newblock Enhancement of bone shadow region using local phase-based ultrasound
  transmission maps.
\newblock International Journal of Computer Assisted Radiology and Surgery
  \textbf{12}(6) (2017)  951--960

\bibitem{batch_normalization}
Ioffe, S., Szegedy, C.:
\newblock Batch normalization: Accelerating deep network training by reducing
  internal covariate shift.
\newblock In: Proceedings of the 32nd International Conference on Machine
  Learning (ICML-15). (2015)  448--456

\bibitem{alexnet}
Krizhevsky, A., Sutskever, I., Hinton, G.E.:
\newblock Imagenet classification with deep convolutional neural networks.
\newblock In Pereira, F., Burges, C.J.C., Bottou, L., Weinberger, K.Q., eds.:
  Advances in Neural Information Processing Systems 25.
\newblock Curran Associates, Inc. (2012)  1097--1105

\bibitem{vgg}
Simonyan, K., Zisserman, A.:
\newblock Very deep convolutional networks for large-scale image recognition.
\newblock arXiv preprint arXiv:1409.1556 (2014)

\bibitem{simultaneous}
Kurmann, T., Neila, P.M., Du, X., Fua, P., Stoyanov, D., Wolf, S., Sznitman,
  R.:
\newblock Simultaneous recognition and pose estimation of instruments in
  minimally invasive surgery.
\newblock In: International Conference on Medical Image Computing and
  Computer-Assisted Intervention, Springer (2017)  505--513

\bibitem{adam_opt}
Kingma, D., Ba, J.:
\newblock Adam: A method for stochastic optimization.
\newblock In: Proceedings of the International Conference on Learning
  Representations (ICLR). (2015)

\end{thebibliography}

\end{document}